\documentclass[10pt,twocolumn,letterpaper]{article}

\usepackage{cvpr}
\usepackage{times}
\usepackage{epsfig}
\usepackage{graphicx}
\usepackage{amsmath}
\usepackage{amssymb}
\usepackage{hhline}
\usepackage[inline]{enumitem}
\usepackage{multirow}
\usepackage[normalem]{ulem}
\usepackage[dvipsnames,table]{xcolor}
\usepackage[pagebackref=true,breaklinks=true,letterpaper=true,colorlinks,bookmarks=false, citecolor=ForestGreen]{hyperref}

\usepackage[square,numbers,sort&compress]{natbib}
\usepackage{algpseudocode}
\usepackage[font=footnotesize,labelfont=bf]{caption}
\usepackage{array}
\usepackage{multirow}
\usepackage{booktabs}
\usepackage{algorithm}
\usepackage{subcaption}
\usepackage[normalem]{ulem}
\usepackage{xparse}
\usepackage{pifont}
\usepackage{bm}
\usepackage{listings}
\usepackage{mwe} \usepackage{makecell}
\usepackage{color, colortbl}
\usepackage{cleveref}

\usepackage[scaled=0.85]{DejaVuSansMono}

\crefformat{section}{\S#2#1#3} \crefformat{subsection}{\S#2#1#3}
\crefformat{subsubsection}{\S#2#1#3}
\crefformat{table}{Table~#2#1#3}
\crefformat{tables}{Tables~#2#1#3}
\crefformat{figure}{Figure~#2#1#3}

\newcommand{\algorithmName}{ClusterFit\xspace}
\newcommand{\algorithmShort}{CF\xspace}
\newcommand{\numClustersNoMath}{K}
\newcommand{\numClusters}{$\numClustersNoMath$\xspace}

\definecolor{pretrainColor}{rgb}{0.5, 0.0, 0.5}
\definecolor{clusterColor}{rgb}{0.0, 0.5, 0.0}
\definecolor{targetColor}{rgb}{0.0, 0.3, 0.8}

\newcommand{\pretrainNetNoMath}{{\color{pretrainColor}\mathrm{N_{pre}} }}
\newcommand{\pretrainNetBold}{{$\color{pretrainColor}\mathbf{N_{pre}} $}}
\newcommand{\pretrainNetLongerNoMath}{{\color{pretrainColor}\mathrm{N_{pre}} }2\times}
\newcommand{\pretrainDataNoMath}{ {\color{pretrainColor}\mathrm{D_{pre}}} }
\newcommand{\pretrainLabelNoMath}{ {\color{pretrainColor}\mathrm{L_{pre}}} }

\newcommand{\pretrainNet}{$\pretrainNetNoMath$\xspace}
\newcommand{\pretrainNetLonger}{$\pretrainNetLongerNoMath$\xspace}
\newcommand{\pretrainData}{$\pretrainDataNoMath$\xspace}
\newcommand{\pretrainLabel}{$\pretrainLabelNoMath$\xspace}

\newcommand{\clusterNetNoMath}{{\color{clusterColor} \mathrm{N_{cf}} }}
\newcommand{\clusterDataNoMath}{ {\color{clusterColor}\mathrm{D_{cf}}} }
\newcommand{\clusterLabelNoMath}{ {\color{clusterColor}\mathrm{L_{cf}}} }

\newcommand{\clusterNet}{$\clusterNetNoMath$\xspace}
\newcommand{\clusterData}{$\clusterDataNoMath$\xspace}
\newcommand{\clusterLabel}{$\clusterLabelNoMath$\xspace}

\newcommand{\targetDataNoMath}{ {\color{targetColor} \mathrm{D_{tar}} }}

\newcommand{\targetAllNoMath}{ {\color{targetColor} \mathrm{T_{tar}} }}
\newcommand{\targetData}{$\targetDataNoMath$\xspace}

\newcommand{\targetAll}{$\targetAllNoMath$\xspace}

\newcommand{\rotationNoSp}{RotNet}
\newcommand{\jigsawNoSp}{Jigsaw}
\newcommand{\rotation}{\rotationNoSp \xspace}
\newcommand{\jigsaw}{\jigsawNoSp \xspace}

\newcommand{\kinetics}{Kinetics\xspace}
\newcommand{\sportsfull}{Sports1M\xspace}

\newcommand{\sthsth}{Something-Something V1\xspace}
\newcommand{\sthsthshort}{Sth-Sth V1\xspace}

\newcommand{\igverburu}{IG-Verb-19M\xspace}

\newcommand{\igvideonoun}{IG-Noun-19M\xspace}
\newcommand{\igkineticsuru}{IG-Kinetics-19M\xspace}
\newcommand{\igverball}{IG-Verb-62M\xspace}
\newcommand{\igimageIN}{IG-ImageNet-1B\xspace}

\newcommand{\ImNetDataset}{ImageNet\xspace}
\newcommand{\ImNet}{ImageNet-1K\xspace}
\newcommand{\ImNetNine}{ImageNet-9K\xspace}

\newcommand{\iNat}{iNaturalist\xspace}

\newcommand{\VOCseven}{VOC07\xspace}
\newcommand{\Placestwo}{Places205\xspace}
\newcommand{\Placesthree}{Places365\xspace}

\newcommand{\resnetfifty}{ResNet-50\xspace}

\newcommand{\convone}{\texttt{conv1}\xspace}
\newcommand{\resfive}{\texttt{res5}\xspace}
\newcommand{\resfour}{\texttt{res4}\xspace}

\newcommand{\videohighcap}{R(2+1)D-34\xspace}
\newcommand{\videolowcap}{R(2+1)D-18\xspace}

\newlength\savewidth\newcommand\shline{\noalign{\global\savewidth\arrayrulewidth
  \global\arrayrulewidth 1pt}\hline\noalign{\global\arrayrulewidth\savewidth}}

\newlength\thinwidth\newcommand\thinline{\noalign{\global\savewidth\arrayrulewidth
  \global\arrayrulewidth 0.5pt}\hline\noalign{\global\arrayrulewidth\savewidth}}

\definecolor{Gray}{gray}{0.93}
\newcolumntype{a}{>{\columncolor{Gray}}c}
\definecolor{LightCyan}{rgb}{0.88,1,1}
\definecolor{highlightRowColor}{gray}{0.9}

\cvprfinalcopy

\begin{document}

\title{ClusterFit: Improving Generalization of Visual Representations}\author{
Xueting Yan\thanks{$^\dagger$ Equal Contribution} \quad \quad Ishan Misra$^*$ \quad \quad Abhinav Gupta \quad \quad Deepti Ghadiyaram$^\dagger$ \quad \quad Dhruv Mahajan$^\dagger$ \\
Facebook AI
}

\maketitle

\begin{abstract}

Pre-training convolutional neural networks with weakly-supervised and self-supervised strategies is becoming increasingly popular for several computer vision tasks. However, due to the lack of strong discriminative signals, these learned representations may overfit to the pre-training objective (e.g., hashtag prediction) and not generalize well to downstream tasks. In this work, we present a simple strategy - \emph{\algorithmName (\algorithmShort)} to improve the robustness of the visual representations learned during pre-training. Given a dataset, we (a) cluster its features extracted from a pre-trained network using k-means and (b) re-train a new network from scratch on this dataset using cluster assignments as pseudo-labels. We empirically show that clustering helps reduce the pre-training task-specific information from the extracted features thereby minimizing overfitting to the same. Our approach is extensible to different pre-training frameworks -- weak- and self-supervised, modalities -- images and videos, and pre-training tasks -- object and action classification. Through extensive transfer learning experiments on $11$ different target datasets of varied vocabularies and granularities, we show that \emph{\algorithmShort} significantly improves the representation quality compared to the state-of-the-art large-scale (millions / billions) weakly-supervised image and video models and self-supervised image models.

\end{abstract}

\section{Introduction}
\label{sec:introduction}

Weak and self-supervised pre-training approaches offer scalability by exploiting free annotation. But \emph{there is no free lunch} -- these methods often first optimize a \textit{proxy objective} function, for example, predicting image hashtags~\cite{joulin2016learning} or color from grayscale images~\cite{larsson2016learning,zhang2016colorful}. Similar to supervised pre-training, the underlying assumption (hope) is that this proxy objective function is fairly well aligned with the subsequent transfer tasks, thus optimizing this function could potentially yield suitable pre-trained visual representations. While this assumption holds mostly true in case of fully-supervised pre-training, it may not extend to weak and self-supervision. In the latter pre-training cases, the lack of strong discriminative signals may result in an undesirable scenario where the visual representations overfit to the idiosyncrasies of the pre-training task and dataset instead, thereby rendering them unsuitable for transfer tasks. For instance, it was noted in~\cite{mahajan2018exploring,sun2017revisiting, ghadiyaram2019large} that factors such as label noise, polysemy (\texttt{apple} the fruit vs. \texttt{Apple Inc.}), linguistic ambiguity, lack of `visual'ness of tags (e.g. \texttt{\#love}) significantly hampered the pre-training proxy objective from being well-aligned with the transfer tasks. Further, the authors of~\cite{zhang2017split, goyal2019scaling} studied multiple self-supervised methods and observed that, compared to earlier layers, features from the last layer are more ``aligned'' with the proxy objective, and thus generalize poorly to target tasks.

In this work, we ask a simple question -- is there a way to avoid such overfitting to the proxy objective during weak- and self-supervised pre-training? Can we overcome the `artifacts' of proxy objectives so that the representation is generic and transferable? 
Our key insight is that \textit{smoothing} the feature space learned via proxy objectives should help us remove these artifacts and avoid overfitting to the the proxy objective. But how do we smoothen the feature space? Should it be done while optimizing the proxy objective or in a post-hoc manner?  
\begin{table}
\centering
\setlength{\tabcolsep}{0.15em}\scalebox{0.78}{
\begin{tabular}{l|l}
\textbf{Pre-training method (\pretrainNet)} & \textbf{$\mathbf{\Delta}$ of \algorithmShort (\clusterNet) on transfer}  \\
\shline
\textbf{Fully-supervised Images}~\cref{sec:synthetic_noise},~\cref{fig:noise}{\color{red}b} & +2.1\% on \ImNetNine~\cite{deng2009imagenet}\\
\resnetfifty, \ImNet, 1K labels & \\

\hline
\rowcolor{highlightRowColor}  \textbf{Weakly-supervised Images}~\cref{sec:ws_image},~\cref{tab:image_ws} & +4.6\% on \ImNetNine~\cite{deng2009imagenet}\\
\rowcolor{highlightRowColor} \small{\resnetfifty, 1B Images, 1.5K hashtags~\cite{mahajan2018exploring}} & +5.8\% on \iNat~\cite{van2018inaturalist}\\

\hline
\textbf{Weakly-supervised Videos}~\cref{sec:ws_video},~\cref{tab:video_ws} & +3.2\% on Kinetics~\cite{kinetics}\\
\small{\videohighcap, 19M videos, 438 hashtags~\cite{ghadiyaram2019large}} & +4.3\% on Sports1M~\cite{sports1m}\\

\hline
\rowcolor{highlightRowColor} \textbf{Self-supervised Images}~\cref{sec:self_sup_images}, Tables~\ref{tab:self-sup}, \ref{tab:self-sup-multi} & +7-9\% on \ImNet~\cite{ILSVRC15}\\
\rowcolor{highlightRowColor}  \small{\resnetfifty, 1M images} & +3-7\% mAP on \VOCseven~\cite{Everingham15} \\
\rowcolor{highlightRowColor}  \small{\jigsaw~\cite{noroozi2016unsupervised} and \rotation~\cite{gidaris2018unsupervised}}, Multi-task (\cref{sec:multi_task_self}) & +3-5\% on \Placestwo~\cite{zhou2014learning}\\
\thinline
\end{tabular}}
\vspace{-0.1in}
\caption{\textbf{A summary of results}: We show that \textbf{\algorithmName (\algorithmShort)} can be applied to a variety of different pre-training methods, modalities, and architectures. We report absolute gains in top-1 accuracy (except for \VOCseven where we report mAP). In each setting, \algorithmShort provides improvements with the \emph{same} model architecture and \emph{without} additional data or supervision.}
\vspace{-0.2in}
\label{tab:highlight_results}
\end{table}

\begin{figure}
    \centering
    \includegraphics[width=0.44\textwidth]{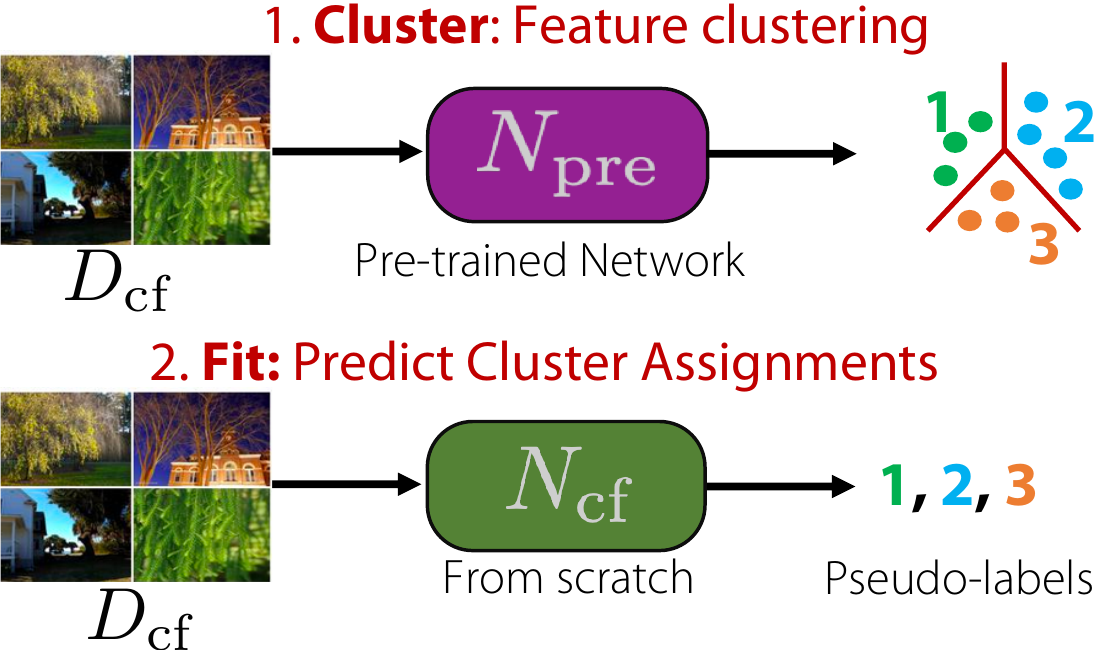}
    \vspace{-0.1in}
    \caption{\footnotesize{\textbf{\algorithmName (\algorithmShort):} We start with a pre-trained network (\pretrainNet) that is trained on some pre-training task (not shown). We use this network to extract features and cluster a new dataset \clusterData using k-means clustering. We show that training a new network \clusterNet from scratch on these cluster assignments as labels results in a more transferable feature representation.}}
    \vspace{-0.2in}
    \label{fig:teaser}
\end{figure}

To this end, we propose a surprisingly simple yet effective framework called \textbf{\algorithmName (\algorithmShort)}. Specifically, given a pre-trained network trained using a proxy objective and a new dataset, we first use the learned feature space to cluster that dataset. Next, we train a new network from scratch on this new dataset using the cluster memberships as pseudo labels (\cref{fig:teaser}). We demonstrate that clustering of the features helps retain only the essential invariances in them and eliminates proxy objective's artifacts (essentially smoothing the feature space). Re-training on the cluster memberships yields a visually coherent pre-training feature space for downstream tasks. Our approach of feature space smoothing
is guided through unsupervised k-means clustering, making it scalable to millions (billions) of videos and images in both weak- and self-supervised pre-training frameworks. 

We take inspiration from recent work in self-supervised learning which aims to learn a smooth visual feature space via clustering and trains representations on the clusters as classes~\cite{caron2018deep,caron2019deep,noroozi2018boosting}. While~\cite{caron2018deep,caron2019deep} use clustering as the training objective itself, in our work, we investigate the value of post-hoc smoothing. \algorithmName can also be viewed as a variant of knowledge distillation~\cite{hinton2015distilling} that distills via `lossy' clustering, as opposed to the standard setup of using soft targets in original label space. 

\algorithmName demonstrates significant performance gains on a total of $11$ public, challenging image and video benchmark datasets. As summarized in~\cref{tab:highlight_results}, our approach, while extremely simple, consistently improves performance across different pre-training methods, input modalities, network architectures, and benchmark datasets.

\section{Related Work}
\label{sec:related_work}

\begin{figure}[!t]
    \centering
    \includegraphics[width=0.5\textwidth]{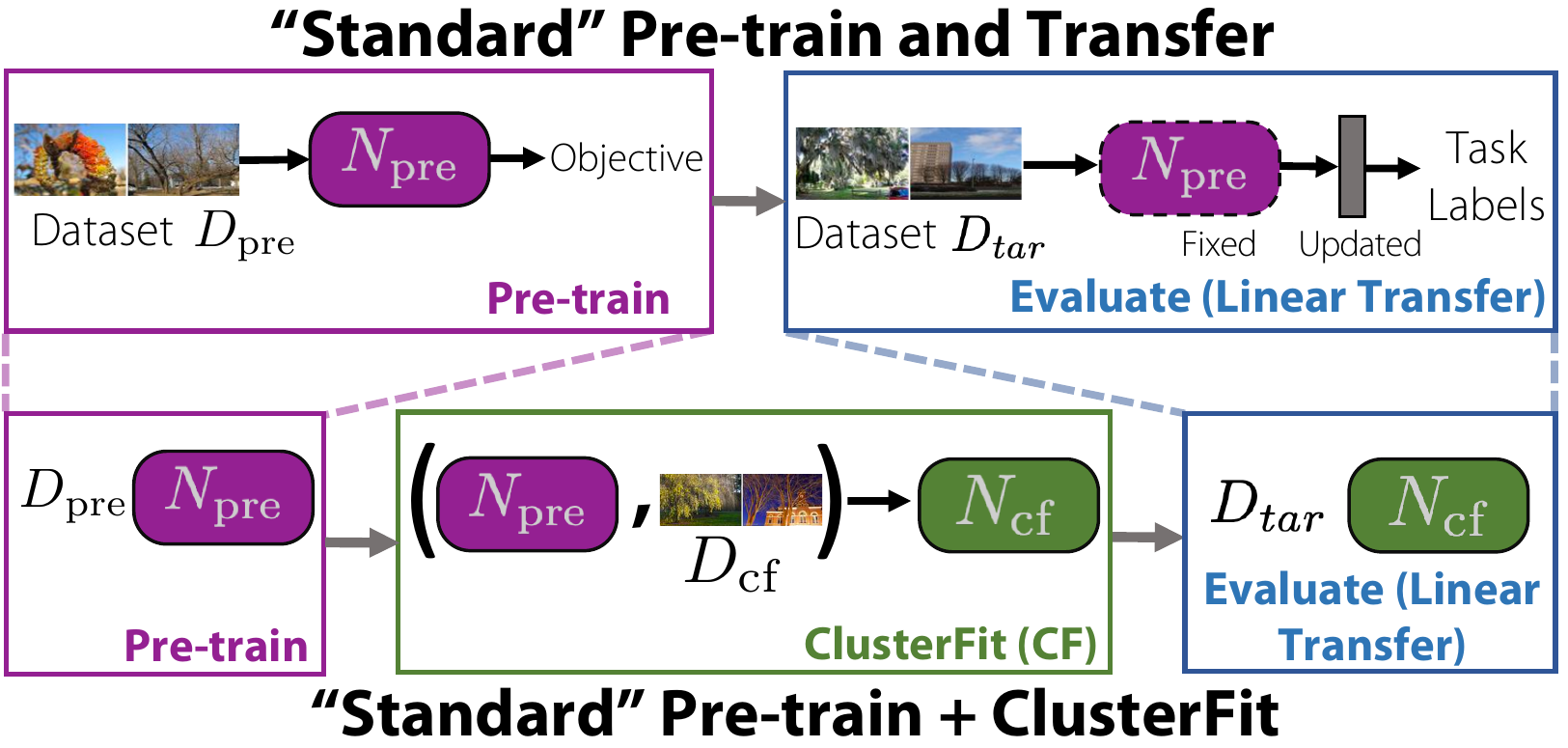}
    \caption{\textbf{Full \algorithmName pipeline:} A typical transfer learning framework involves two stages: pre-training followed by transfer learning. \algorithmName introduces a step between these stages. We evaluate all representations by training a linear classifier on \emph{fixed} ConvNet weights.}
    \vspace{-0.3in}
    \label{fig:pipeline}
\end{figure}

\par \noindent \textbf{Weakly Supervised Learning:} Training ConvNets on very large, weakly supervised images by defining the proxy tasks using the associated meta-data~\cite{mahajan2018exploring, sun2017revisiting, gross2017hard, joulin2016learning, li2017visualngram, denton2015hashtag, taigman2015webscale, schroff2015facenet, thomee2016yfcc100m, veit2017separating, ghadiyaram2019large} has shown tremendous benefits. Proxy tasks include hashtags predictions~\cite{mahajan2018exploring, veit2017separating, denton2015hashtag, gross2017hard, ghadiyaram2019large}, GPS~\cite{hays2008im2gps,vo2017revisiting}, search queries prediction~\cite{sun2017revisiting}, and word or n-grams predictions~\cite{joulin2016learning, li2017visualngram}.
Our approach builds upon these works and shows that even better representations can be trained by leveraging the features from such pre-training frameworks for clustering to mitigate the effect of noise. Yalniz \etal~\cite{zeki2019semi} propose a target task specific noise removal framework by ranking images for each class by their softmax values and retaining only top-$K$ images for re-training. However, their method is specific to a particular target task and discards most of the data during re-training. By contrast, our approach does not adhere to a particular target task and leverages all the data, since, they may contain complementary visual information beyond hashtags.

\par \noindent \textbf{Self-Supervised Learning:} Self-supervised approaches typically learn a feature representation by defining a `pre-text' task on the visual domain. These pre-text tasks can either be domain agnostic~\cite{bojanowski2017unsupervised,oord2018representation,wu2018unsupervised,hjelm2018learning,caron2018deep,xie2016unsupervised} or exploit domain-specific information like spatial structure in images~\cite{doersch2015unsupervised,noroozi2016unsupervised, noroozi2017representation,noroozi2018boosting,gidaris2018unsupervised}, color~\cite{deshpande2015learning,larsson2016learning,larsson2017colorization,zhang2017split,zhang2016colorful}, illumination~\cite{dosovitskiy2016discriminative}, temporal structure~\cite{mobahi2009deep,hadsell2006dimensionality,fernando2017self,misra2016shuffle,luc2017predicting} or a co-occurring modality like sound~\cite{arandjelovic2017look,arandjelovic2018objects,gao2018learning,owens2016ambient,de1994learning}. In this work, we use two diverse image-based self-supervision approaches - \jigsaw~\cite{noroozi2016unsupervised} and \rotation~\cite{gidaris2018unsupervised} that have shown competitive performance~\cite{goyal2019scaling,caron2019deep,kolesnikov2019revisiting}. Since the difference between pretext tasks and semantic transfer learning tasks is huge, our method shows much larger improvement for self-supervised methods (\cref{sec:self_sup_images}).

Our work builds upon~\cite{caron2018deep,caron2019deep}, who use clustering and pseudo-labels for self-supervised learning and~\cite{noroozi2018boosting}, who
distill predictions from different self-supervised models to a common architecture. Compared to~\cite{caron2018deep,caron2019deep}, \algorithmName does not require any alternate optimization and thus is more stable and computationally efficient. As we show in~\cref{experiments_all}, this property makes \algorithmName easily scalable to different modalities and large-scale data. Compared to~\cite{noroozi2018boosting}, our focus is not distilling information to a common architecture, but instead to remove the pre-training task biases. This makes \algorithmName applicable broadly to any kind of pre-trained models - fully supervised or use noisy supervision (\cref{sec:synthetic_noise}), weakly supervised from billions of images or millions of videos (\cref{sec:weak_sup}), and self-supervised models (\cref{sec:self_sup_images}).

\par \noindent \textbf{Model Distillation:} Model distillation~\cite{ba2014deep,hinton2015distilling,furlanello2018born,anil2018large} typically involves transferring knowledge from a `teacher' model to a `student' model by training the student on predictions of the teacher in addition to task labels. These methods are designed to transfer knowledge (not contained in the labels) about the task from the teacher to the student network. Since distillation retains more knowledge about the original task, it performs poorly in the case of weak-supervision (\cref{sec:weak_sup}). Interestingly, the failure of standard knowledge distillation approaches in the context of self-supervised learning has also been shown in ~\cite{noroozi2018boosting}. 

\begin{table}[!]
~\centering
\setlength{\tabcolsep}{0.15em}\scalebox{0.7}{
\begin{tabular}{lcccc}
\textbf{Dataset} & \textbf{Label Type} & \textbf{\# classes} & \textbf{Train/Eval} & \textbf{Metric}\\
\shline
\multicolumn{5}{c}{Weakly-supervised Images~\cref{sec:ws_image}}\\
\thinline
\textbf{\ImNet}~\cite{ILSVRC15} & multi-class object & 1000  & 1.3M/50K& top-1 acc\\
\rowcolor{highlightRowColor} \textbf{\ImNetNine}~\cite{deng2009imagenet} & multi-class object & 9000 & 10.5M/450K& top-1 acc\\
\textbf{\Placesthree}~\cite{zhou2014learning} & multi-class scene & 365  & 1.8M/36.5K& top-1 acc\\
\rowcolor{highlightRowColor}  \textbf{\iNat 2018}~\cite{van2018inaturalist} & multi-class object & 8142  & 438K/24K& top-1 acc\\
\thinline
\multicolumn{5}{c}{Weakly-supervised Videos~\cref{sec:ws_video}}\\
\thinline
\textbf{\kinetics}~\cite{kinetics} & multi-class action & 400 & 246K/20K & top-1 acc \\
\rowcolor{highlightRowColor} \textbf{\sportsfull}~\cite{sports1m} & multi-class action & 487 & 882K/204K & top-1 acc\\
\small{\textbf{\sthsth}}~\cite{sth-sth} & multi-class action & 174  & 86K/11.5K & top-1 acc\\
\thinline
\multicolumn{5}{c}{Self-supervised Images~\cref{sec:self_sup_images}}\\
\thinline
\textbf{\VOCseven}~\cite{Everingham15} & multi-label object & 20 & 5K/5K& mAP\\
\rowcolor{highlightRowColor}  \textbf{\ImNet}~\cite{ILSVRC15} & multi-class object & 1000  & 1.3M/50K& top-1 acc\\
\textbf{\Placestwo}~\cite{zhou2014learning} & multi-class scene  & 205 & 2.4M/21K& top-1 acc\\
\rowcolor{highlightRowColor} \textbf{\iNat 2018}~\cite{van2018inaturalist} & multi-class object & 8142  & 438K/24K& top-1 acc\\
\hline
\end{tabular}
}
\vspace{-0.1in}
~\caption{\textbf{Target tasks for Transfer Learning} used for evaluating feature representations.}
\vspace{-0.2in}
\label{tab:target_data}
\end{table}

\begin{figure*}[!t]
\centering
\includegraphics[width=0.9\linewidth]{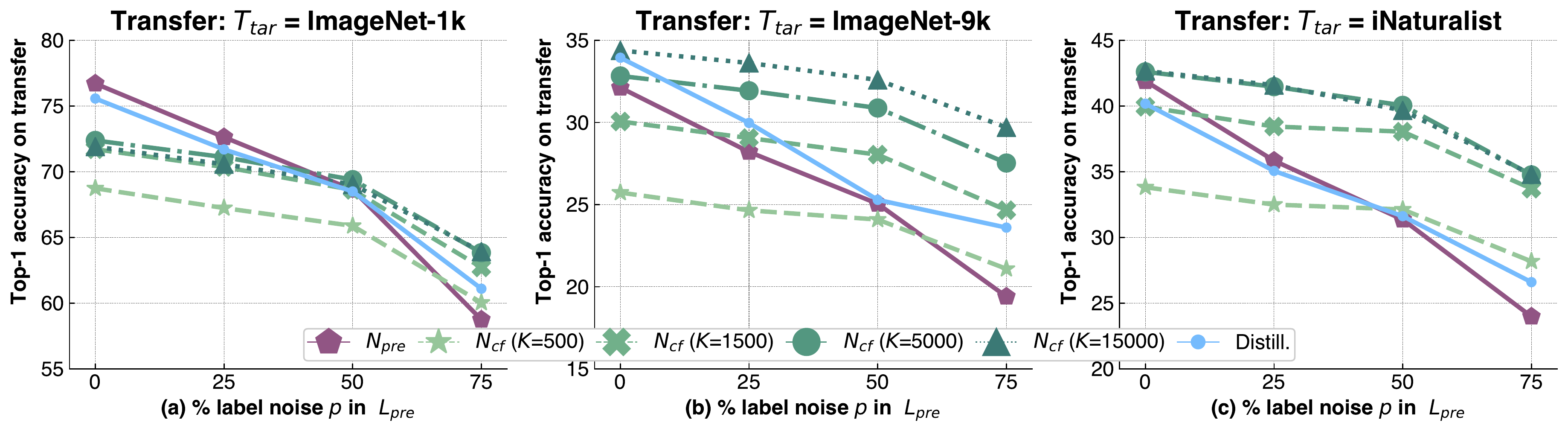}
\vspace{-0.15in}
\caption {\textbf{Control Experiment:} We inject uniform label noise in the labels \pretrainLabel from  $\pretrainDataNoMath\!=\xspace$ \ImNet and train a separate \resnetfifty model (\pretrainNet) on these noisy labels. We apply \algorithmName on each of these pre-trained models (\pretrainNet) and vary the number of clusters \numClusters to train \clusterNet. We then study the transfer learning performance of the representations by training a linear classifier on fixed features from \pretrainNet or \clusterNet on three target tasks - a noise free \ImNet, \ImNetNine, and \iNat. \algorithmName is able to learn more transferable features despite high amounts of label noise in pre-training. For finer-grained target tasks like \ImNetNine, \algorithmName can even improve a fully supervised \resnetfifty model ($p\!=\!0$).}
\label{fig:noise}
\end{figure*}

\section{Approach} \label{sec:approach}
Our goal is to learn a generalizable feature space for a variety of target tasks that does not overfit to the pre-training proxy objective. We first describe the framework of \algorithmName (\algorithmShort) in \cref{sec:formulation}. Next, we report a control experiment on the \ImNet dataset that sheds light on how \algorithmShort combats the `bias' introduced due to the proxy objective (\cref{sec:synthetic_noise}).

\subsection{\algorithmName Framework} \label{sec:formulation}

Our method starts with a ConvNet \pretrainNet that is pre-trained on a dataset \pretrainData and labels \pretrainLabel. 
First, we use the penultimate layer of \pretrainNet to extract features from each datapoint belonging to another dataset \clusterData. Next, we \textbf{cluster} these features using k-means into \numClusters groups and treat these cluster assignments as the new categorical `labels' (\clusterLabel) for \clusterData. Finally, we \textbf{fit} a different network \clusterNet (initialized from scratch) on \clusterData that minimizes a cross-entropy objective on \clusterLabel. We illustrate these steps in Figure~\ref{fig:teaser}. We highlight that re-learning \clusterNet from scratch on \clusterData is completely unsupervised and thus allows leveraging large-scale datasets.

\par \noindent \textbf{Intuition:} We hypothesize that \algorithmName (\algorithmShort) leverages the underlying visual smoothness in the feature space to create visually coherent clusters. We believe that ``cluster'' followed by ``fit'' weakens the underlying pre-training objective-specific bias. One may view \algorithmName from an information bottleneck~\cite{tishby2015deep} perspective wherein the `lossy' clustering step introduces a bottleneck and removes any pre-training proxy objective bias.

\par \noindent \textbf{How to evaluate \algorithmShort?} As in prior efforts~\cite{mahajan2018exploring,goyal2019scaling,ghadiyaram2019large}, we use transfer learning performance on downstream tasks to understand whether \algorithmShort improves generalization of the feature representations. Specifically, to evaluate \pretrainNet and \clusterNet, we train linear classifiers on \emph{fixed} feature representations from the networks on the downstream task 
and report final performance on held-out data (see~\cref{tab:target_data}). Figure~\ref{fig:pipeline} illustrates \algorithmName's setup.
We stress that \algorithmName is simple to implement and makes minimal assumptions about input modalities, architectures \etc but provides a powerful way to improve the generalization of the feature space. We explore various design choices \algorithmName offers such as relative properties of \pretrainNet, \clusterNet, \pretrainData, and \clusterData in~\cref{sec:ablation_all}.

\subsection{Control Experiment using Synthetic Noise}
\label{sec:synthetic_noise}
Here, our goal is to study the extent of generalization of features learned from a `proxy' pre-training objective in a controlled setup. We start with a supervised pre-training dataset \ImNet~\cite{ILSVRC15}, and add synthetic label noise to it. Our motive behind this setup is to intentionally misalign the pre-training objective with downstream tasks. We acknowledge that the synthetic noise simulated in this experiment is an over simplification of the complex noise present in real world data. Nevertheless, it provides several key insights into \algorithmName as we show next.

\par \noindent \textbf{Control Experiment Setup:} To isolate the effect of \algorithmShort, in this experiment, we fix $\pretrainDataNoMath\!=\!\clusterDataNoMath\!=\xspace$ \ImNet and the network architectures \pretrainNet and \clusterNet to \resnetfifty~\cite{he2016deep}. We start by adding varying amounts ($p\%$) of uniform random label noise\footnote{We randomly replace a label ($l$) in \ImNet \texttt{train} split with one that is obtained by uniformly sampling from \ImNet labels excluding $l$.} to \pretrainData.
Next, we train a separate $\pretrainNetNoMath$ for each fraction $p$ of the noisy labels. We then apply \algorithmShort (with different values of \numClusters in k-means) to each $\pretrainNetNoMath$ to obtain a corresponding $\clusterNetNoMath$. Finally, we evaluate the representations by training linear classifiers on fixed \resfive features on three target image classification datasets - \ImNet, \ImNetNine, and \iNat.
We use model distillation~\cite{hinton2015distilling} as a baseline to better understand the behavior of \algorithmName.

Our motivation behind this setup is the following: when $p\!=\!0$, \pretrainNet denotes the true, noise-free supervised task; as $p$ increases, the proxy objective becomes a poorer approximation of the original pre-training objective and allows us to closely inspect \algorithmName.

\par \noindent \textbf{Results and Observations:} We report the transfer learning performance of $\pretrainNetNoMath$ (i.e., before \algorithmShort) and $\clusterNetNoMath$ (i.e., after \algorithmShort) in \cref{fig:noise} for different values of label noise $p$. Let us first consider $p=0$, \ie, a setting without any label noise. In this case, \pretrainNet is trained on clean labels. On the target dataset \ImNet, $\pretrainNetNoMath$ performs significantly better than $\clusterNetNoMath$ for all values of $K$ (Fig.~\ref{fig:noise} (a)). This is expected, since when $\pretrainDataNoMath\!=\!\targetDataNoMath=\xspace$ \ImNet, the pre-training and transfer tasks are exactly aligned. However, $\clusterNetNoMath$ performs comparably or better than $\pretrainNetNoMath$ for other target da - \ImNetNine and \iNat at higher values of $K$. This suggests that \algorithmShort can \textbf{improve even fully-supervised} representations for more fine-grained downstream tasks. We note that model distillation also provides an improvement over \pretrainNet on \ImNetNine but is worse on \iNat.

Let us now consider scenarios where $p > 0$. Figure \ref{fig:noise} indicates that increased label noise ($p$) in \pretrainData translates to poor performance across all three target tasks. We highlight that the drop in the performance is more drastic for $\pretrainNetNoMath$ (i.e., before \algorithmShort), than for $\clusterNetNoMath$ (i.e., after \algorithmShort). More importantly, the performance gap between $\clusterNetNoMath$ and $\pretrainNetNoMath$ continues to increase with $p$. From Fig.~\ref{fig:noise} (b) and (c), we observe that $\clusterNetNoMath$ consistently outperforms $\pretrainNetNoMath$ on two target tasks \ImNetNine and \iNat. Notably, for $\targetDataNoMath\!=\xspace$ \ImNet (\cref{fig:noise} (a)), when $p\geq50$, $\clusterNetNoMath$ outperform $\pretrainNetNoMath$, which is pre-trained on noisy \ImNet. Model distillation provides some gains over \pretrainNet but is consistently outperformed by \algorithmName.

These results suggest that as $p$ increases, the proxy objective gets further away from the `true' pre-training objective, and makes features from $\pretrainNetNoMath$ less transferable. In those very cases, \algorithmShort captures useful visual invariances in the feature representations, thereby providing more noise-resilient pseudo-labels for learning transferable representations. Finally, we also note that larger number of clusters \numClusters generally leads to better transfer learning performance. The gains are larger for more challenging and fine-grained datasets like \ImNetNine and \iNat. We study the effect of this hyper-parameter \numClusters in~\cref{sec:ablation_all}.

\section{Experiments} 
\label{experiments_all}
We now examine the broad applicability of \algorithmName in three different pre-training scenarios for \pretrainNet: (a) weakly-supervised pre-training for images~ (\cref{sec:ws_image}), (b) weakly-supervised pre-training for videos~(\cref{sec:ws_video}), and (c) self-supervised pre-training for images~(\cref{sec:self_sup_images}). 

\par \noindent \textbf{Common \algorithmShort Setting:} Throughout this section, we set \pretrainData $=$ \clusterData and \pretrainNet $=$ \clusterNet (architecture-wise). We train \pretrainNet on \pretrainData, \clusterNet on \clusterData for equal number of epochs. ~\cref{tab:arch_and_data_settings} summarizes these settings. By keeping the data, architecture, and training schedule constant, we hope to measure the difference in performance between \clusterNet and \pretrainNet solely due to \algorithmName. 
\par \noindent \textbf{Evaluation:} As mentioned in~\cref{sec:formulation}, we evaluate \algorithmName via transfer learning on target tasks. Specifically, we train linear classifiers on the \emph{fixed} features obtained from the penultimate layer of \pretrainNet or \clusterNet on target datasets. The transfer learning tasks are summarized in~\cref{tab:target_data}.

\noindent{\textbf{Baselines:}} We use the following baselines:
\vspace{-0.09in}
\begin{itemize}[leftmargin=*,noitemsep]
    \item \noindent \pretrainNetBold: We use features from \pretrainNet for transfer learning. Since \algorithmName (\algorithmShort) is applied on \pretrainNet to get \clusterNet, this baseline serves to show improvements through \algorithmShort.
    
    \item \noindent{\textbf{Distillation:}} To empirically understand the importance of the clustering step in \algorithmShort, we compare with model distillation~\cite{hinton2015distilling}. Unlike \algorithmShort, distillation transfers knowledge from \pretrainNet \textit{without clustering}, thus retaining more information about the learned features. We train a distilled model using a weighted average of $2$ loss functions: (a) cross-entropy with soft targets computed using \pretrainNet and temperature $T$ and (b) cross-entropy with image/video labels in weakly-supervised setup. We also experimented with training a network to directly regress the features from \pretrainNet but found consistently worse results.
    
    \item \noindent{\textbf{Prototype:}} \algorithmName uses unsupervised k-means to create pseudo-labels. To understand the effect of this unsupervised step, we add a baseline that uses \textit{semantic information} during clustering. Under this prototype alignment~\cite{prototypical} baseline, unlike random cluster initialization as done in k-means, we use label information in \clusterData to initialize cluster centers. Specifically, we first set \numClusters equal to the number of `classes' in \clusterData. Here, each cluster corresponds to a `prototype' of that class. We then compute \numClusters prototypes by averaging image embeddings of all images belonging to each class. Finally, pseudo-labels are assigned to each data point by finding its nearest `prototype' cluster center. Since this method uses explicit label information present in \clusterData, it requires more `supervision' than \algorithmName. We also note that this baseline is not applicable to self-supervised methods (suppl. material).        
    \item \noindent \textbf{Longer pre-training:} Since \clusterNet is trained for the same number of epochs as \pretrainNet, we also compare against a network trained on the pre-train task for $2\times$ longer (denoted by \pretrainNetLonger). Specifically, \pretrainNetLonger is trained for a combined number of epochs as \pretrainNet and \clusterNet. By comparing \clusterNet against this baseline, we hope to isolate improvements due to longer pre-training.
    \end{itemize}
\begin{table}[!]
\footnotesize
~\centering
\setlength{\tabcolsep}{0.2em}\scalebox{0.9}{

\begin{tabular}{l|c|c}
\textbf{Pre-training method} & $\pretrainDataNoMath\!=\! \clusterDataNoMath$ & \textbf{Arch. of \pretrainNet \& \clusterNet} \\
\shline
Weakly-Supervised Images~\cref{sec:ws_image} & \igimageIN & \resnetfifty  \\
\rowcolor{highlightRowColor} Weakly-Supervised Videos~\cref{sec:ws_video} & \igverburu & \videohighcap \\
Self-supervised Images~\cref{sec:self_sup_images} & ImageNet-1k & \resnetfifty \\
\thinline
\end{tabular}
}
\vspace{-0.1in}
~\caption{\textbf{Data and model architectures used} in~\cref{experiments_all}: weakly supervised videos, weakly supervised images, and self supervised images. In each setting, we train \pretrainNet and \clusterNet for equal number of epochs.}
\label{tab:arch_and_data_settings}
\end{table}
\subsection{Weakly-supervised pre-training}
\label{sec:weak_sup}
In this section, we study weakly-supervised pre-training on noisy web images and videos. These approaches predict the noisy hashtags associated with images/videos and thus minimize a \emph{proxy objective} during pre-training.
\begin{table}[!]
\centering
\setlength{\tabcolsep}{0.25em}\scalebox{0.75}{
\begin{tabular}{l|c|c|c|c|ccccc}

\multirow{2}{*}{\targetData} & \multirow{2}{*}{\pretrainNet} & \multirow{2}{*}{\pretrainNetLonger} & \multirow{2}{*}{\textbf{Distill.}} & \multirow{2}{*}{\textbf{Prototype}} & \multicolumn{5}{c}{\textbf{\algorithmShort (\clusterNet)}, \numClusters $\rightarrow$}\\ 
 & & & & & $1.5k$ & $3.75k$ & $7.5k$ & $15k$ & $30k$\\ 

\shline
\ImNet & 78.0 & \textbf{78.8} & 73.8 & 76.9 & 75.3 & 76.1 & \underline{76.5} & \underline{76.5} & 76.2\\ \rowcolor{highlightRowColor} \ImNetNine & 32.9 & 34.1 & 29.1 & 35.1 & 33.5 & 35.4 & 36.4 & 37.1 & \textbf{37.5}\\ \Placesthree & \underline{51.2} & 51.2 & 49.9 & 51.9 & 52.0 & 52.1 & 52.4 & \textbf{52.6} & 52.1\\ \rowcolor{highlightRowColor} \iNat & 43.9 & 45.3 & 35.9 & 49.0 & 43.8 & 46.4 & 47.9 & \textbf{49.7} & 49.5\\
\hline
\end{tabular}
}
\vspace{-0.1in}
\caption { \textbf{Weakly-supervised Images:} Top-1 accuracy for various transfer learning datasets with \pretrainData = \clusterData = \igimageIN and the same architecture (\resnetfifty) for \pretrainNet and \clusterNet.}
\vspace{-0.2in}
\label{tab:image_ws}
\end{table}

\subsubsection{Weakly-supervised image pre-training} 
\label{sec:ws_image}
\par \noindent  \textbf{Data and Model:} As in~\cite{mahajan2018exploring}, we collect \igimageIN dataset of 1B public images associated with hashtags from a social media website. To construct this dataset, we consider images tagged with at least one hashtag that maps to any of the \ImNet synsets. The architecture of \pretrainNet and \clusterNet network is fixed to a \resnetfifty~\cite{he2016deep}, while \pretrainData $=$ \clusterData $=$ \igimageIN. 
\par \noindent  \textbf{\algorithmName Details:} We extract features from the $2048$ dimensional \resfive layer from \pretrainNet for clustering. \clusterNet is trained from scratch on $\clusterDataNoMath$ $=$ \igimageIN on the cluster assignments as pseudo-labels. Details on the hyper parameters during pre-training and \algorithmName are provided in the supplementary material. We report results in~\cref{tab:image_ws}, which we discuss next.

\par \noindent  \textbf{Effect of longer pre-training:} \pretrainNet pre-trained on \pretrainData $= $\igimageIN already exhibits very strong performance on all target datasets. By construction, the label space of the target dataset \ImNet matches with that of \pretrainData. As noted in~\cite{mahajan2018exploring}, this translates to \pretrainNet yielding an impressive top-1 accuracy of 78\% on \ImNet. Features from longer pre-training (\pretrainNetLonger) show improvements on \ImNet, \ImNetNine, and \iNat but not on \Placesthree. As noted in~\cite{joulin2016learning,mahajan2018exploring}, \Placesthree is not well-aligned with \ImNet (and by extension with \igimageIN). Thus, (longer) pre-training yields no benefit. By contrast, the target dataset  \ImNetNine is well-aligned with \pretrainData $=$ \igimageIN, thus achieving improvements from longer pre-training.

\par \noindent \textbf{Comparison with Model Distillation:} Training a student network via distillation, \ie, soft targets provided by the teacher (\pretrainNet) and hashtags, performs worse than \pretrainNet itself. In our case, the student and teacher network are of the same capacity (\resnetfifty). We believe that the noisy label setting combined with the same capacity student and teacher networks are not ideal for model distillation.

\par \noindent \textbf{Comparison with Prototype:} Except on \ImNet, the prototype baseline shows improvement over both \pretrainNet and \pretrainNetLonger. This shows that pseudo-labels derived based on label information can provide a better training objective than hashtags used for pre-training \pretrainNet. However, similar to \algorithmShort, prototype shows a reduction in performance on \ImNet which we explain next.

\par \noindent  \textbf{Gains of \algorithmName:} \clusterNet achieves substantial gains over the strong \pretrainNet model especially on fine-grained datasets like \ImNetNine (\textbf{4.6} points) and \iNat (\textbf{5.8} points), at higher values of $K$. This may be because \clusterNet captures a more diverse and finer-grained visual feature space that benefits fine-grained transfer tasks. We observe a small decrease in the performance on \ImNet (1.5 points) which can be attributed again to the hand-crafted label alignment of the \igimageIN with \ImNet. This result is inline with observations from~\cite{mahajan2018exploring}. We believe the performance decrease of `prototype' on \ImNet is also due to this reason. \clusterNet shows improved performance than `prototype,' yet does not use any additional supervision while generating pseudo-labels. Finally, we note that finding an optimal number of clusters $K$ for each transfer learning task is procedurally easier than finding a pre-training task (or label space) that aligns with the target task.

\begin{table}[!]
\centering
\setlength{\tabcolsep}{0.3em}\scalebox{0.75}{
\begin{tabular}{l|c|c|c|c|ccccc}

\multirow{2}{*}{\targetAll} & \multirow{2}{*}{\pretrainNet} & \multirow{2}{*}{\pretrainNetLonger} & \multirow{2}{*}{\textbf{Distill.}} & \multirow{2}{*}{\textbf{Prototype}} & \multicolumn{5}{c}{\textbf{\algorithmShort (\clusterNet), \numClusters $\rightarrow$}}\\ 
& & & & & $400$ & $800$ & $1600$ & $3200$ & $6400$\\ 

\shline
\kinetics & 68.8 & 69.2 & 63.6 & 70.3 & 70.1 & 71.2 & 71.2 & 71.5 & \textbf{72.0}\\ \rowcolor{highlightRowColor} \sportsfull & 52.9 & 53.1 & 48.4 & 55.1 & 55.8 & 56.6 & 57.1 & \textbf{57.2} & 57.2\\ \sthsthshort & 16.9 & 16.4 & 15.6 & 20.3 & 20.2 & 20.0 & \textbf{20.6} & 19.3 & 19.7\\ \hline
\end{tabular}
}
\vspace{-0.1in}
\caption {\textbf{Weakly-supervised videos:}  Top-1 accuracy for various transfer learning datasets with \pretrainData = \clusterData = \igverburu and the same architecture (\videohighcap) for \pretrainNet and \clusterNet. 
}
\vspace{-0.2in}
\label{tab:video_ws}
\end{table}

\subsubsection{Weakly-supervised video pre-training}\label{sec:ws_video}
\par \noindent \textbf{Data and Model:} Following~\cite{ghadiyaram2019large}, we collect \igverburu, a dataset of $19M$ public videos with hashtags from a social media website. We consider videos tagged with at least one of the $438$ verbs from Kinetics~\cite{kinetics} and VerbNet~\cite{verbnet}. We set \pretrainData $=$ \clusterData $=$ \igverburu. We use the clip-based R(2+1)D-34~\cite{r2plus1D} architecture for \pretrainNet and \clusterNet. Each video clip is generated by scaling its shortest edge to $128$ followed by cropping a random patch of size $112\!\times\!112$. We use $32$ consecutive frames per video clip, with temporal jittering applied to the input.
\par \noindent \textbf{\algorithmName details:} We uniformly sample $6$ clips of $32$ consecutive frames per video, extract video features per clip, and average pool them. We use the $512$ dimensional \resfive layer from \pretrainNet. We direct the reader to the supplementary material for hyper-parameter details.

\par \noindent  \textbf{Observations:} We present the transfer learning results in Table~\ref{tab:video_ws}. Once again, the baseline \pretrainNet exhibits strong performance on all target datasets. Longer pretraining (\pretrainNetLonger) provides limited benefit on
\kinetics and \sportsfull, and loses performance compared to \pretrainNet on \sthsthshort. As observed in~\cref{sec:ws_image}, model distillation performs worse than \pretrainNet on all target datasets.

We observe that \algorithmShort (\clusterNet) provides significant improvements of $\textbf{3.2 - 4.3}\%$ across all the datasets over \pretrainNet. The optimal number of clusters \numClusters vary depending on each dataset, but is typically an order of magnitude higher than the size of the original label space (i.e., $438$ verbs in \igverburu). For example, performance does not saturate for \kinetics even at $K=6400$. We study the effect of $K$ in~\cref{sec:clustering}.

\begin{figure*}[!t]
\centering
\includegraphics[width=0.75\linewidth]{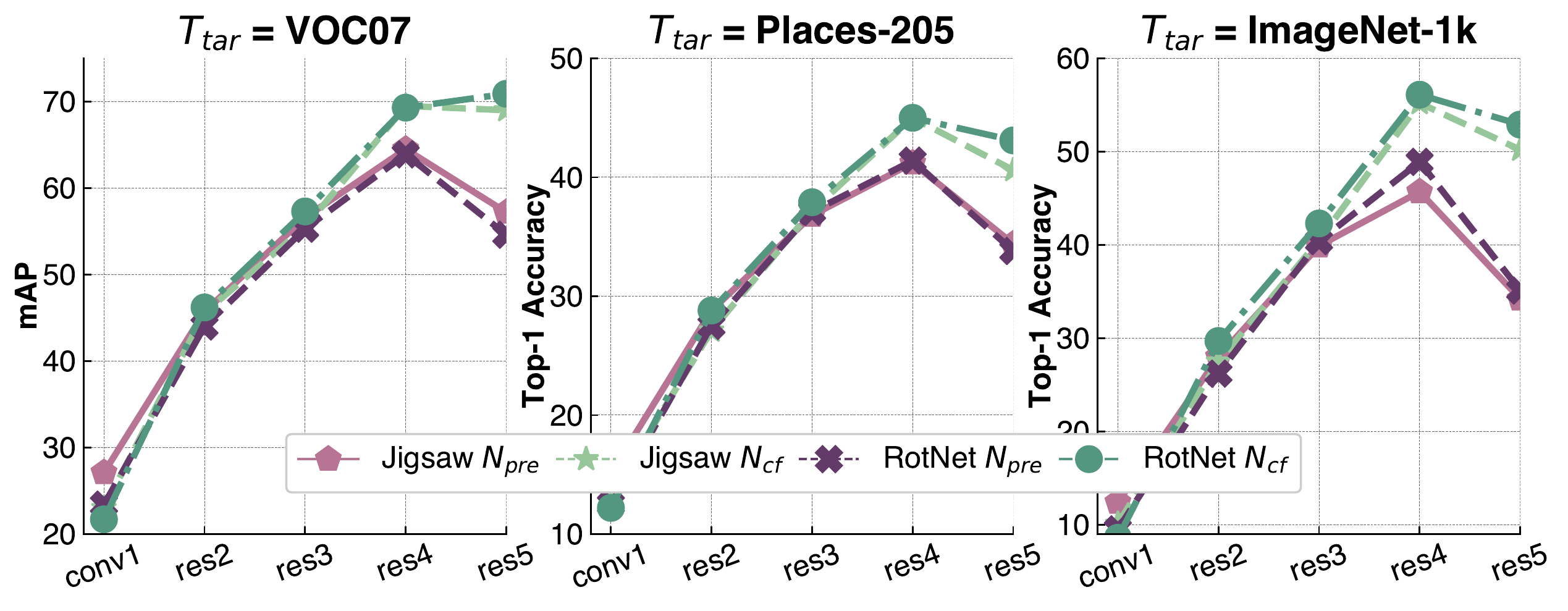}
\vspace{-0.1in}
\caption {\textbf{Self-supervised Images (Layerwise):} We examine the layer-wise performance of self-supervised models before applying our technique (\pretrainNet) and after (\clusterNet). We fix $\pretrainDataNoMath\!=\!\clusterDataNoMath\!=\ $\ImNet (without labels) and use the same architecture (\resnetfifty) for \pretrainNet and \clusterNet. The last layer (\resfive) features for \pretrainNet transfer poorly compared to the lower \resfour layer. After \algorithmShort, \clusterNet shows an improved performance for all layers except for \convone and reduces the gap in performance between \resfour and \resfive.}
\vspace{-0.2in}
\label{fig:selfsup_layerwise}
\end{figure*}

\subsection{Self-Supervised pre-training for Images}
\label{sec:self_sup_images}

We now apply \algorithmName framework to self-supervised methods.
We study two popular and diverse self-supervised methods - \jigsawNoSp~\cite{noroozi2016unsupervised} and \rotationNoSp~\cite{gidaris2018unsupervised}. These methods do not use semantic labels and instead create pre-training labels using a `pre-text' task such as rotation. As mentioned in~\cref{sec:related_work} and~\cite{noroozi2018boosting}, distillation is not a valid baseline for these self-supervised methods (more in supplementary material). Also, as these methods do not use semantic label information, `prototype' is also not a valid baseline.
\par \noindent \textbf{Data and Model:} We fix the network architectures of \pretrainNet and \clusterNet to \resnetfifty. 
We also fix \pretrainData $=$ \clusterData = \ImNet
to pre-train \jigsaw and \rotation models (\pretrainNet). We discard the semantic labels and use only images from both tasks. We use the models released by~\cite{goyal2019scaling} for \jigsaw and train \rotation models following the approach in~\cite{gidaris2018unsupervised,goyal2019scaling}.
\par \noindent \textbf{\algorithmName Details:} We set $\numClustersNoMath \!=\! 16,000$. \clusterNet is trained for the same number of epochs as the pre-trained self-supervised network \pretrainNet.
We strictly follow the training hyper parameters and the transfer learning setup outlined in Goyal \etal~\cite{goyal2019scaling}. We report additional results for different values of \numClusters in the supplemental material.

\par \noindent \textbf{Layer-wise transfer:} In~\cref{fig:selfsup_layerwise}, we report the transfer learning performance of each layer of \pretrainNet and compare with \clusterNet after applying \algorithmName. We see that for the pre-trained network \pretrainNet, \texttt{res5} features transfer poorly compared to \texttt{res4} features. For example, on \VOCseven dataset, linear classifiers trained on \texttt{res4} perform $\sim$ 3-10 points better than those trained on \texttt{res5} for both \jigsaw and \rotation networks. As noted in~\cite{zhang2017split, goyal2019scaling}, this is because the final layer features overfit to the pre-training (`pre-text') task. 

After applying \algorithmName, we see that features of \clusterNet transfer better across all the layers except for \convone -- an improvement of \textbf{7 to 9 points} on \ImNet -- for both \jigsaw and \rotation methods. On \VOCseven, \texttt{res5} features transfer better than \texttt{res4}: for \pretrainNet the gap is $-9$ points while for \clusterNet it is about $+1$ points. On \ImNet and \Placestwo, the performance gap of \clusterNet when using \texttt{res4} vs. \texttt{res5} features is considerably reduced. 
This strongly suggests that \algorithmName reduces the overfitting of \texttt{res5} features to the pre-text task, thus making them generalize better.

\par \noindent \textbf{Results:} We show additional transfer learning results in~\cref{tab:self-sup}.
Longer pre-training (\pretrainNetLonger) shows mixed results -- a small drop in performance for \jigsaw and a small increase in performance  for \rotation. \algorithmName provides consistent improvements on both \jigsaw and \rotation tasks, across all pre-training and target tasks. We achieve significant boosts of \textbf{3-5 points} on \Placestwo and \textbf{5-8 points} on \iNat.

\par \noindent \textbf{Easy multi-task Learning using \algorithmName:} In~\cref{sec:multi_task_self}, we show that \algorithmName can be easily applied to combine multiple different self-supervised methods and provides impressive gains of more than \textbf{8 points} on \ImNet in top-1 accuracy.

\begin{table}[!]
\centering
\setlength{\tabcolsep}{0.2em}\scalebox{0.8}{
\begin{tabular}{l|acac}
\hspace{0.3in} \textbf{Method} & \multicolumn{4}{c}{\targetAll}\\
 & \textbf{\ImNet} & \textbf{\VOCseven} & \textbf{\Placestwo} & \textbf{\iNat}\\
\shline

\jigsaw \pretrainNet & 46.0 & 66.1 & 39.9 & 22.1\\
\jigsaw \pretrainNetLonger & 45.1 & 65.4 &  38.7 & 21.8 \\
\jigsaw \clusterNet (Ours) & \textbf{55.2} & \textbf{69.5} & \textbf{45.0} & \textbf{29.8}\\
\hline
\rotation \pretrainNet & 48.9 & 63.9 & 41.4 & 23.0\\
\rotation \pretrainNetLonger  & 50.0 & 64.9 & 42.9 & 25.3\\
\rotation \clusterNet (Ours) & \textbf{56.1} & \textbf{70.9} & \textbf{44.8} & \textbf{28.4}\\
\thinline

\end{tabular}}
\caption{\textbf{Self-supervised methods:} We apply \algorithmName to self-supervised methods and evaluate them following the setup in~\cite{goyal2019scaling} on four datasets by training a linear classifier on fixed features. All methods use the \resnetfifty architecture for \pretrainNet and \clusterNet. We report the performance of the best performing layer for each method and use the mean Average Precision (mAP) metric for the \VOCseven dataset and top-1 accuracy for all other datasets.}
\label{tab:self-sup}
\end{table}

\vspace{0.1in}
\par \noindent \textbf{Summary:} We demonstrate that the misalignment between pre-training and transfer tasks due to the high levels of noise in the web data or the non-semantic nature of the self-supervised pretext tasks leads to a less-generalizable feature space. Through extensive experiments, we show that \algorithmName consistently combats this issue across different modalities and pre-training settings.
\section{Analyzing \algorithmName} 
\label{sec:ablation_all}
\algorithmName involves several aspects such as the relative model capacities of \pretrainNet and \clusterNet, properties of \pretrainData and \clusterData, size and granularity of the pre-training label space, and so on. In this section, we study the effect of these design choices on the transfer learning performance with videos as an example use case (Table \ref{tab:target_data}).\\
\noindent \textbf{Experimental Setup:} Similar to \igverburu in Sec.~\ref{sec:ws_video}, we construct \igverball, a weakly-supervised dataset comprising $62M$ videos and use it as \clusterData. For faster training of \clusterNet, we consider a computationally cheaper  \videolowcap~\cite{r2plus1D} architecture and process $8$ frames per video clip. Unless specified otherwise, \pretrainData $=$ \igverburu and \pretrainNet $=$ \videohighcap~\cite{r2plus1D} with $32$ frames per video. All other settings are same as in Sec.~\ref{sec:ws_video}.

\begin{figure}
    \centering
    \includegraphics[width=0.4\textwidth]{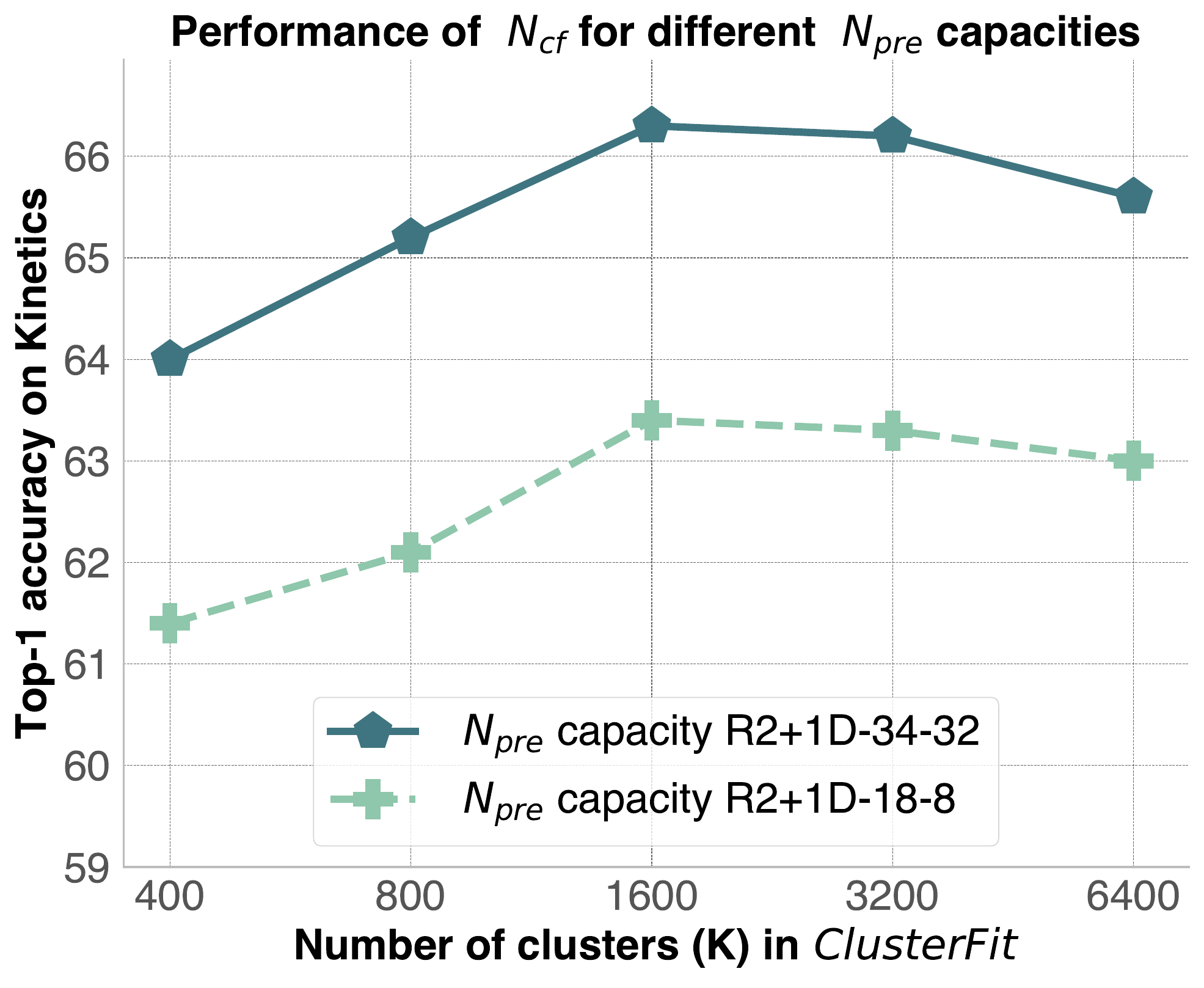}
    \caption{\textbf{Relative Model Capacity of \pretrainNet and \clusterNet (\cref{sec:feature_model_capacity}):} We fix \clusterNet $=$ \videolowcap. We vary (a) \pretrainNet $=$ \clusterNet $=$ \videolowcap (light green) and (b) \pretrainNet $>$ \clusterNet, where \pretrainNet $=$ \videohighcap (dark green). We report the transfer performance of the \clusterNet model for cases (a) and (b) on \kinetics. A higher capacity \pretrainNet results in better transfer performance.}
    \vspace{-0.2in}
     \label{fig:model_capacity}
\end{figure}
\subsection{Relative model capacity of \pretrainNet and \clusterNet} 
\label{sec:feature_model_capacity}
The relative model capacities of \pretrainNet and \clusterNet can impact the final transfer performance of \clusterNet. To study this behavior, we fix \pretrainData $=$ \igverburu and \clusterData $=$ \igverball, and \clusterNet $=$ \videolowcap. We vary the architecture of \pretrainNet as follows: (a) \pretrainNet $=$ \clusterNet $=$ \videolowcap ; (b) \pretrainNet $>$ \clusterNet, where \pretrainNet $=$ \videohighcap model ($64M$ parameters) and thus higher capacity than \clusterNet ($33M$ parameters). 

From~\cref{fig:model_capacity}, we observe a consistent improvement of $2\%-3\%$ across different values of \numClusters when a higher capacity model was used as \pretrainNet. This result is intuitive and indicates that a higher capacity \pretrainNet yields richer visual features for clustering and thus improves the transfer learning performance. We note that in the aforementioned case (b), our framework can be viewed to be distilling knowledge from a higher capacity teacher model (\pretrainNet) to a lower-capacity student model (\clusterNet).
\subsection{Unsupervised vs. Per-Label Clustering}
\label{sec:clustering}
\begin{table}[!]
\centering
\setlength{\tabcolsep}{0.1em}
\begin{tabular}{c}
\includegraphics[width=0.5\textwidth]{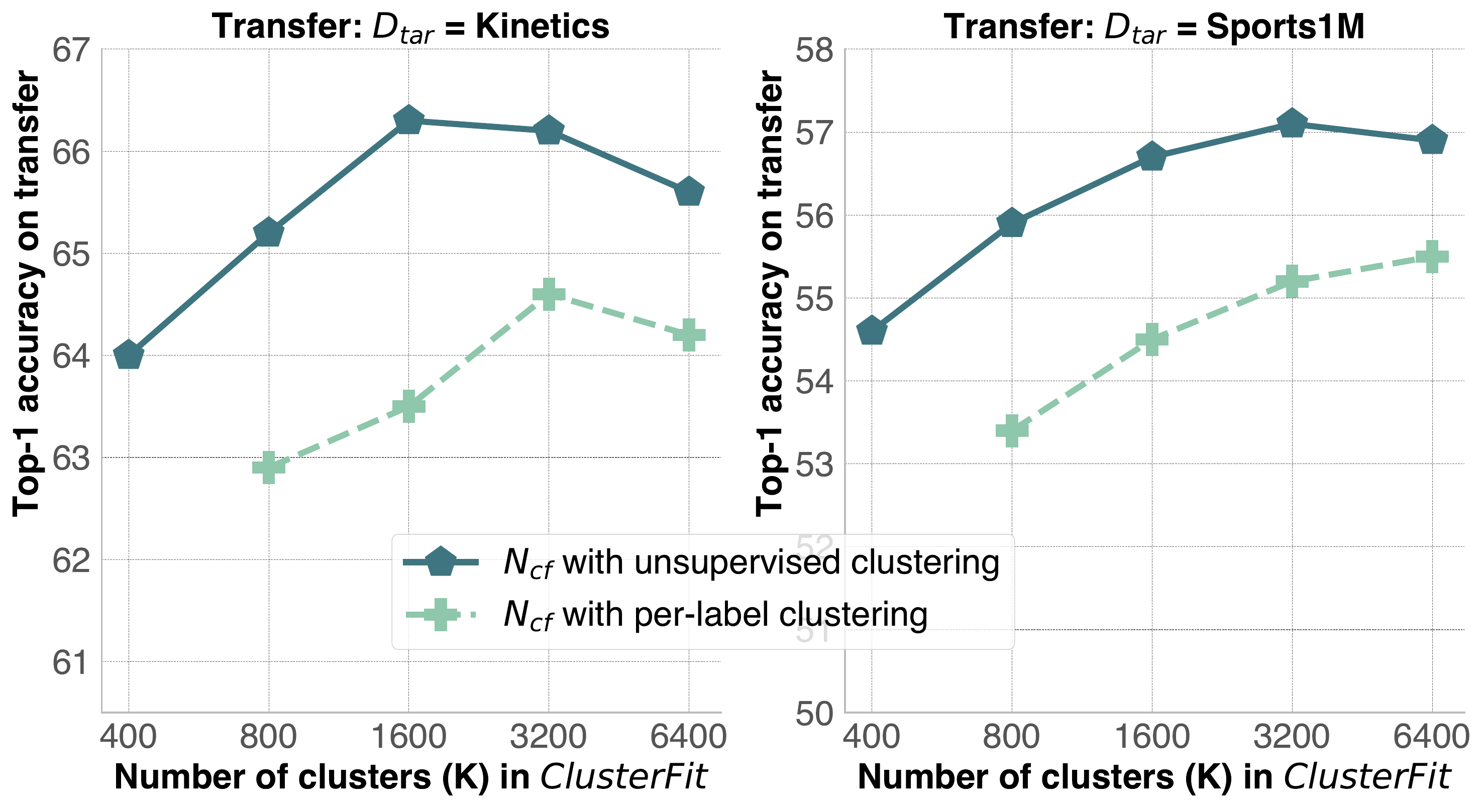}
\end{tabular}
\captionof{figure}{\textbf{Unsupervised vs. Per-Label Clustering~(\cref{sec:clustering}):} 
In per-label clustering, we retain the semantic information of the class labels and cluster videos belonging to each label. We note that for all values of $K$, unsupervised clustering used in \algorithmName yields better transfer learning performance on \kinetics and \sportsfull.}
\vspace{-0.2in}
\label{fig:cluster_figure}
\end{table}
As noted before, the clustering step in \algorithmName is `unsupervised' because it discards the labels associated with \clusterData and operates purely on the feature representations. But is there any advantage of using the semantic information of labels in \clusterData for clustering? To address this question, we formulate a per-label clustering setup. Specifically, given each label $l$, we cluster videos belonging to it into $k_l$ clusters. We treat $K = \{k_l  : \forall l \}$ as pseudo-labels to train \clusterNet. Each $k_l$ is defined to be proportional to $\sqrt{n_l}$
\footnote{We also experimented with $k_l \approx {n_l}$ but this resulted in worse performance.} where $n_l$ denotes the number of videos associated with the label $l$. 

Figure~\ref{fig:cluster_figure} compares the two clustering approaches on \kinetics and \sportsfull. We observe that on both datasets, unsupervised clustering consistently outperforms per-label clustering across all values of $K$. We believe that by operating purely on video features, the unsupervised approach effectively captures the visual coherence in \clusterData. Consequently, factors around label noise such as wrong / missing labels and lexical ambiguity are being automatically addressed in the unsupervised framework, leading to superior performance over per-label clustering.

\subsection{Properties of \pretrainData} \label{sec:type_of_pretrain_data}
In this section, we address the following question: what constitutes a valuable pre-training label space (\pretrainData) and how to construct one? Towards this end, we study two properties of \pretrainData : the nature of it's labels and their cardinality. We refer the readers to the supplementary material for discussion on the nature of labels.

\begin{figure}[t]
    \centering
    \includegraphics[width=.4\textwidth]{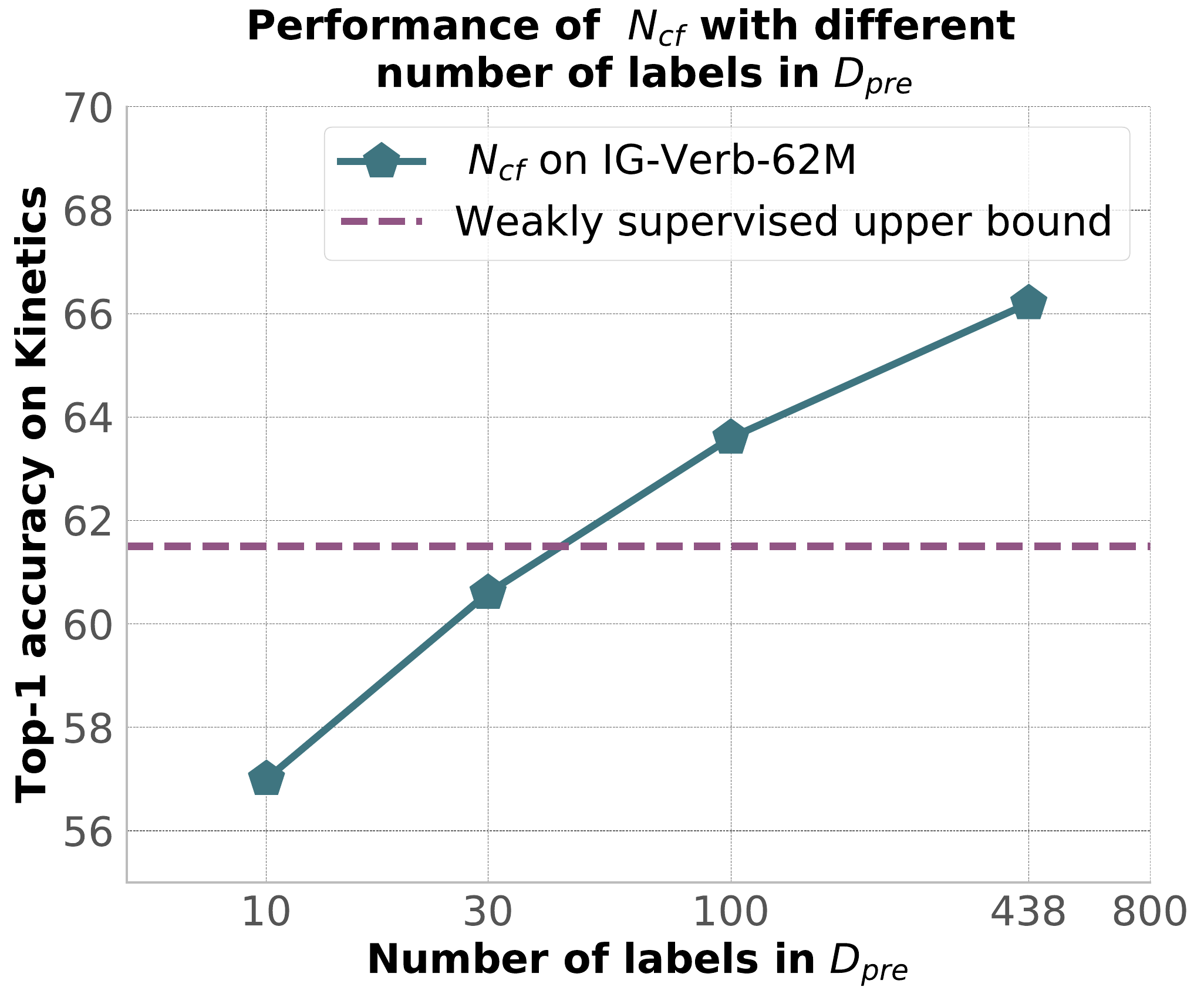}
    \caption{\textbf{Effect of number of labels in \pretrainNet (~\cref{sec:type_of_pretrain_data})}. We design $4$ different \pretrainData, each with $19M$ videos but \#labels $ = \{10,30,100,438\}$, $K = 3200$ and \clusterData $=$ \igverball. X-axis is in log-linear scale.}
    \label{fig:num_labels}
    \vspace{-0.2in}
\end{figure}

\noindent \textbf{Number of labels in \pretrainData:}
We now study how varying the number of labels in \pretrainData effects \algorithmName. To study this, we fix the total number of unique videos in \pretrainData to $19M$ and vary the number of pre-training labels. First, we consider \igverball and rank it's $438$ weak verb labels by their frequency of occurrence. Next, we construct $4$ different \pretrainData datasets by considering unique $19M$ videos tagged with top-m verbs, where $m = \{10, 30, 100, 438\}$. Note that for a fixed number of videos in \pretrainData, reducing the number of labels implies reduced content diversity. 

From Figure~\ref{fig:num_labels}, we observe that the transfer learning performance increases log-linearly with the number of pre-training labels in \pretrainData. When we use just the top-10 verbs ($m = 10$), accuracy drops by around $9\%$ compared to $m = 438$. This indicates that label space diversity is essential to generate good quality clusters. However, when $m = 100$, $\clusterNetNoMath$ is within $2\%$ of the accuracy obtained when using all $438$ verbs, and it outperform its weakly supervised pre-trained counterpart which uses 62M videos and all 438 verbs.  This experiment clearly demonstrates the utility of our approach in designing a generic pre-training label space with minimal effort. Contrary to~\cite{mahajan2018exploring, ghadiyaram2019large} which propose careful, manual label engineering, \algorithmName offers an easy way to construct a powerful, generalizable pre-training label space. Increasing the label space granularity is as simple as increasing the number of clusters in \algorithmName and requires no additional manual effort.

\section{Discussion}
In this work, we presented  \algorithmName, a simple approach to significantly improve the generalizability of features learnt in weakly-supervised and self-supervised frameworks for images and videos. While models trained in these frameworks are prone to overfit to the pre-training objective, \algorithmName combats this issue by first clustering the original feature space and re-learning a new model on cluster assignments. Clustering in \algorithmShort may be viewed as a \textit{lossy} compression scheme that effectively captures the essential visual invariances in the feature space. Thus, predicting the cluster labels gives the `re-learned' network an opportunity to learn features that are less sensitive to the original pre-training objective, making them more transferable.

While the clustering step in \algorithmName is unsupervised, in practice, domain knowledge from downstream target tasks can be used to guide clustering and possibly improve the transfer learning performance. Additionally, we found that in its current unsupervised form, iterative application of \algorithmShort provides little improvements; incorporating domain knowledge could be a potential solution.

\algorithmName is a universal framework - it is scalable and imposes no restrictions on model architectures, modalities of data, and forms of supervision. Future research should take advantage of its flexibility and combine different types of pre-trained models for learning cluster assignments in a multi-task manner. Using evidence accumulation methods~\cite{strehl2002cluster,nguyen2007consensus,fred2002data} for clustering is another worthwhile direction to explore.

{\small
\paragraph{Acknowledgments:} We would like to thank Rob Fergus, and Zhenheng Yang for feedback on the manuscript; Filip Radenovic, and Vignesh Ramanathan for feedback on the experimental setup; Laurens van der Maaten, Larry Zitnick, Armand Joulin, and Xinlei Chen for helpful discussions.}

{\small
\bibliographystyle{ieee_fullname}
\bibliography{refs}
}

\clearpage
\appendix
\section*{Supplemental Material}

\section{Details on baseline methods}
 \paragraph{Details on distillation} The distillation loss function is a convex combination of two individual losses - (1) a loss that tries to match the `soft' target outputs by a teacher model, \ie, a probability distribution computed by applying a softmax function on the logits of a teacher model using a temperature $T$; (2) a cross-entropy loss that tries to match the predictions with the ground truth labels for each datapoint. The two losses are combined with a convex combination ($\alpha$ used for the first loss and $1-\alpha$ used for the second loss). We performed a grid search to find the optimal values for temperature $T$ and weight $\alpha$. We set these values as $T=20$ and $\alpha=0.75$. 
  
 \paragraph{Details on prototype baseline}
  
 Prototype method is a simplified version of k-means with only one iteration and controlled initialization. Similar to k-means, it consists of two steps: (1) cluster center initialization and (2) cluster label reassignment. In step 1, we compute an average of the visual features of all datapoints (videos or images) belonging to each label. These visual embeddings per label are chosen as cluster centers. Then, in step 2, data points are re-assigned to their nearest cluster centers (computed from step 1). Finally, the newly assigned cluster id is used as the label in training \clusterNet. Prototype method is limited to (weakly) supervised setting as it requires label information in step 1. Also, it yields the same number of clusters as the cardinality of the input label space.

\section{Weakly supervised Images}
We provide details for Section 4.1.1 of the main paper.

\paragraph{Details on the IG-ImageNet dataset} We get $1500 (> 1000)$ total hashtags because multiple hashtags may map to the same ImageNet synset.

\paragraph{Details on pre-training} We follow the implementation from~\cite{mahajan2018exploring}. All models are trained using Synchronized stochastic gradient descent (SGD) on $128$ GPUs across $16$ machines. Each GPU processes $32$ images at a time. All the pre-training runs process $2B$ images in total. An initial learning rate of $1.6$ is used and decreased by a factor of $2$ at $13$ equally spaced steps.

\paragraph{Details on \algorithmName}
k-means: Since number of images are very large, we randomly sub-sample $200M$ images and perform $30$ iterations of k-means clustering on them. We use the cluster centers obtained in this step, and then perform $5$ more iterations of k-means with all the 1B images.

\paragraph{Transfer Learning Hyperparameters} We closely follow transfer learning settings in \cite{mahajan2018exploring} and apply parameter sweep on learning rate for different target datasets.

\section{Weakly supervised Videos}
We provide details for Section 4.1.2 of the main paper.
\paragraph{Details on the IG-Verb dataset}
We strictly follow \cite{ghadiyaram2019large} to construct large-scale weakly supervised video datasets. We borrow label space from public datasets and crawl videos that contain matching hashtags from a social website. Given the great amount of noise in web data, labels with at least 50 matching videos are finally retained. IG-Verb-62M dataset consists of 438 verbs, a union of Kinetics and VerbNet \cite{verbnet} verbs, and 62M videos with at least one matching hashtag. If multiple hashtags are attached to one video, one hashtag/verb is randomly picked as label. We also follow the tail-preserving strategy in \cite{ghadiyaram2019large} to construct IG-verb-19M, which is a subset of IG-Verb-62M.

\paragraph{Details on the IG-Kinetics and IG-Noun dataset}
Following the same rules as above, we construct IG-Kinetics-19M comprising $359$ labels from Kinetics label space, and IG-Noun-19M comprising $1438$ labels from \ImNetDataset synsets. These two datasets are used later in studying the effect of nature of labels in \pretrainData.

\paragraph{Details on pre-training} Training for both \pretrainNet and \clusterNet follow the same setting. $128$ GPUs across $16$ machines are used. Each GPU processes $6$ videos at a time and batch normalization~\cite{ioffe2015batch} is applied to all convolutional layers on each GPU. All the pre-training experiments process $490M$ videos in total across all epochs. We closely follow the training hyper-parameters mentioned in \cite{ghadiyaram2019large}. 

\paragraph{Transfer Learning Hyperparameters} We closely follow transfer learning settings in \cite{ghadiyaram2019large} and apply parameter sweep on learning rate for different target datasets.

\section{Self-supervised Images}
We provide details for Section 4.2 of the main paper.

\paragraph{Self-supervised Pre-training (\pretrainNet)} The \jigsaw and \rotation model pre-training is based on the code release from~\cite{goyal2019scaling,kolesnikov2019revisiting,gidaris2018unsupervised}. We use a standard \resnetfifty model for both methods. We use a batchsize of $32$ per GPU, a total of 8 GPUs and optimize these models using mini-batch SGD for a total of $105$ epochs with an initial learning rate of $0.1$, decayed by a factor of $10$ after every $30$ epochs.
\paragraph{Details on \jigsaw \pretrainNet} We follow~\cite{goyal2019scaling,noroozi2016unsupervised} to construct the `jigsaw puzzles' from the images. We first resize the image (maintaining aspect ratio) to make its shortest side $255$, and then extract a random square crop of $255\times 255$ from it. This crop is divided into a $3\times3$ grid and a random crop of $64\times64$ is extracted from each of the 9 grids to get 9 patches. The patches are input individually to the network to obtain their features, and are concatenated in a random order. Finally, the concatenated features are input to a classification layer which predicts the `class index' of the random permutation used to concatenate the features. We use $2000$ permutations as used in~\cite{goyal2019scaling}.

\paragraph{Details on \rotation \pretrainNet} We follow~\cite{gidaris2018unsupervised,kolesnikov2019revisiting} and apply a random rotation from $\{0^{\circ}, 90^{\circ}, 180^{\circ}, 270^{\circ}\}$ to the input image. The network is trained (4-way classification) to predict the index of the rotation applied to the input.

\paragraph{Details on \algorithmName} We extract features (\resfive after average pooling, 2048 dimensional vector) from each of the self-supervised \pretrainNet networks on the \ImNet dataset (train split of 1.28M images). We then $l_2$ normalize the features and use k-means to cluster these images and obtain pseudo-labels as the cluster assignments for each point. 

We also trained a version of \clusterNet by clustering the \resfour features from \pretrainNet. We found that this version gave similar performance to the \clusterNet trained on cluster assignments from the \resfive layer of \pretrainNet.

\paragraph{Transfer Learning Hyperparameters} We train linear classifiers on fixed features. Following~\cite{goyal2019scaling} we use mini-batch SGD with a batchsize of $256$, learning rate of $0.01$ dropped by a factor of 10 after two equally spaced intervals, momentum of $0.9$, and weight decay of $5\times10^{-4}$. The features from each layer (\convone, \resfour, \resfive \etc) are average pooled to get a feature of about $9000$ dimensions each. We try to keep the number of parameter updates for training the linear models are roughly constant across all the transfer datasets. Thus, the models are trained for $28$ epochs on \ImNet (1.28M training images), $14$ epochs on \Placestwo (2.4M training images) and for $84$ epochs on iNaturalist-2018 (437K training images). We follow~\cite{goyal2019scaling} and train linear SVMs for the \VOCseven transfer task.

\paragraph{Transfer Learning Results} In Section 4.2 of the main paper, we showed transfer learning results when $\pretrainDataNoMath = \clusterDataNoMath = $ \ImNet and the architecture of \pretrainNet and \clusterNet was \resnetfifty. In Table~\ref{tab:self-sup-numclust} we show results for different values of the number of cluster, \numClusters, used to generate the pseudo-labels. Although the performance of \clusterNet increases as the number of clusters \numClusters increases, we observe that \algorithmName provides significant improvements over the pre-trained \pretrainNet at smaller values ($\numClustersNoMath=1000$).

\begin{table}[!]
\centering
\setlength{\tabcolsep}{0.2em}\scalebox{0.8}{
\begin{tabular}{l|acac}
\hspace{0.3in} \textbf{Method} & \multicolumn{4}{c}{\targetAll}\\
 & \textbf{\ImNet} & \textbf{\VOCseven} & \textbf{\Placestwo} & \textbf{\iNat}\\
\shline

\jigsaw \pretrainNet & 46.0 & 66.1 & 39.9 & 22.1\\
\thinline
\jigsaw \clusterNet ($K=1000$) & 50.2 & 66.2 & 42.5 & 24.5\\
\jigsaw \clusterNet ($K=4000$) & 51.6 & 67.8 & 42.4 & 27.2\\
\jigsaw \clusterNet ($K=16000$) & \textbf{55.2} & \textbf{69.5} & \textbf{45.0} & \textbf{29.8}\\
\shline
\rotation \pretrainNet & 48.9 & 63.9 & 41.4 & 23.0\\
\thinline
\rotation \clusterNet ($K=1000$) & 51.8 & 67.2 & 42.6 & 25.2\\
\rotation \clusterNet ($K=4000$) & 52.3 & 67.4 & 43.4 & 26.8\\
\rotation \clusterNet ($K=16000$) & \textbf{56.1} & \textbf{70.9} & \textbf{44.8} & \textbf{28.4}\\
\shline
\end{tabular}}
\caption{\textbf{Self-supervised methods:} We apply \algorithmName to self-supervised methods and evaluate them following the setup in~\cite{goyal2019scaling} on four datasets by training a linear classifier on fixed features. All methods use the \resnetfifty architecture for \pretrainNet and \clusterNet. We report the performance of the best performing layer for each method and use the mean Average Precision (mAP) metric for the \VOCseven dataset and top-1 accuracy for all other datasets. We show results for different values of \numClusters used to generate the pseudo-labels.}
\label{tab:self-sup-numclust}
\end{table}

\paragraph{Distillation is not a valid baseline} In self-supervised methods like \jigsaw and \rotation, the predictions depend upon the image transformation (permutation of patches for \jigsaw or the rotations for \rotation) applied to the input. Thus, given an untransformed input image, the self-supervised methods do not produce a `distribution' over the possible set of output values. For example, in \rotation, if we only pass an untransformed image, the network predicts (with high confidence) that the image is rotated by $0^\circ$, and thus the `distribution' over the possible rotation values of the input is not very meaningful to use in a distillation method. For \jigsaw, since the output is the index of the permutation applied to the input patches, the distribution produced for a `regular' input is not meaningful.

\section{Bonus: Multi-task Self-supervised Learning}
\label{sec:multi_task_self}
We study the generalization of \algorithmName by using it for self-supervised multi-task learning. We take a pre-trained network \pretrainNet on \jigsaw and use it to compute the pseudo-labels (via clustering) on \clusterData. We repeat the process for another \pretrainNet trained on \rotation to get a different set of pseudo-labels on \clusterData. We treat these two sets of pseudo-labels as two different multi-class classification problems and train a new \clusterNet from scratch using these labels. Thus, \clusterNet is trained with two different fully-connected layers, each of which predicts the pseudo-labels from a different \pretrainNet. We sum the losses from these two layers and optimize the network. 

We follow the setup from~\cref{sec:self_sup_images}, and use the \resnetfifty architecture for both \pretrainNet and \clusterNet, and set $\pretrainDataNoMath\!=\!\clusterDataNoMath\!=\!$\ImNet. The linear evaluation results are presented in Table~\ref{tab:self-sup-multi}. This n{\"a}ive way of multi-task learning using \algorithmName still improves the performance and provides gains of \textbf{8 points on \ImNet}, \iNat in top-1 accuracy and \VOCseven mAP compared to the \pretrainNet models. The multi-task models also improve over the single task \clusterNet models.

\begin{table}[!]
\centering
\setlength{\tabcolsep}{0.2em}\scalebox{0.8}{
\begin{tabular}{l|acac}
\hspace{0.3in} \textbf{Method} & \multicolumn{4}{c}{\targetAll}\\
 & \textbf{\ImNet} & \textbf{\VOCseven} & \textbf{\Placestwo} & \textbf{\iNat}\\
\shline

\jigsaw \pretrainNet & 46.0 & 66.1 & 39.9 & 22.1\\
\jigsaw \pretrainNetLonger & 45.1 & 65.4 &  38.7 & 21.8 \\
\jigsaw \clusterNet (Ours) & 55.2 & 69.5 & 45.0 & 29.8\\
\hline
\rotation \pretrainNet & 48.9 & 63.9 & 41.4 & 23.0\\
\rotation \pretrainNetLonger  & 50.0 & 64.9 & 42.9 & 25.3\\
\rotation \clusterNet (Ours) & 56.1 & 70.9 & 44.8 & 28.4\\
\hline
\jigsaw + \rotation \clusterNet & \textbf{57.0} & \textbf{72.8} & \textbf{46.2} & \textbf{31.6}\\
 ~~~{\small{(Ours, Multi-task)}} \\

\thinline
\end{tabular}}
\caption{\textbf{Multi-task Self-supervised:} We show that \algorithmName can be used for easy multi-task learning. We apply \algorithmName to self-supervised methods and evaluate them following the setup in~\cite{goyal2019scaling} on four datasets by training a linear classifier on fixed features. All methods use the \resnetfifty architecture for \pretrainNet and \clusterNet. We report the performance of the best performing layer for each method and use the mean Average Precision (mAP) metric for the \VOCseven dataset and top-1 accuracy for all other datasets.}
\label{tab:self-sup-multi}
\end{table}

\section{Analysis of ClusterFit} \label{sec:analysis}
We provide details for Section 5 of the main paper.

\paragraph{Details on \clusterNet training:} In Section 5, we use \videolowcap as \clusterNet and \igverball as \clusterData. Training is done with $64$ GPUs across $8$ machines. Each GPU processes $16$ videos at a time and batch normalization\cite{ioffe2015batch} is applied to all convolutional layers on each GPU. The training processes 250M video in total. An initial learning rate of 0.005 per GPU is applied and decreased by a factor of 2 at 13 equally spaced steps.

\paragraph{Effect of nature of labels in \pretrainData:}
In the main paper, we have shown results where \pretrainNet is pre-trained on labels that are \textit{verbs} (i.e., on \igverburu dataset). It is natural to question: how does the choice of \pretrainData's label space effect \algorithmName? To study this, we vary \pretrainData by changing the properties of their labels, but keeping the volume of the data fixed. All other settings, including \clusterData, \pretrainNet and \clusterNet, are fixed. Specifically, we consider three \clusterData datasets with $19M$ videos each: (a) \igverburu, (b) \igvideonoun and (c) \igkineticsuru. Next, we pre-trained three separate \pretrainNet on these datasets. Then, we use \clusterData = \igverball to apply \algorithmName on each of them, and get three \clusterNet. Finally, transfer learning is done on three \clusterNet.

\begin{table}[!]
\centering
\setlength{\tabcolsep}{0.1em}
\begin{tabular}{cc}
\includegraphics[width=0.24\textwidth]{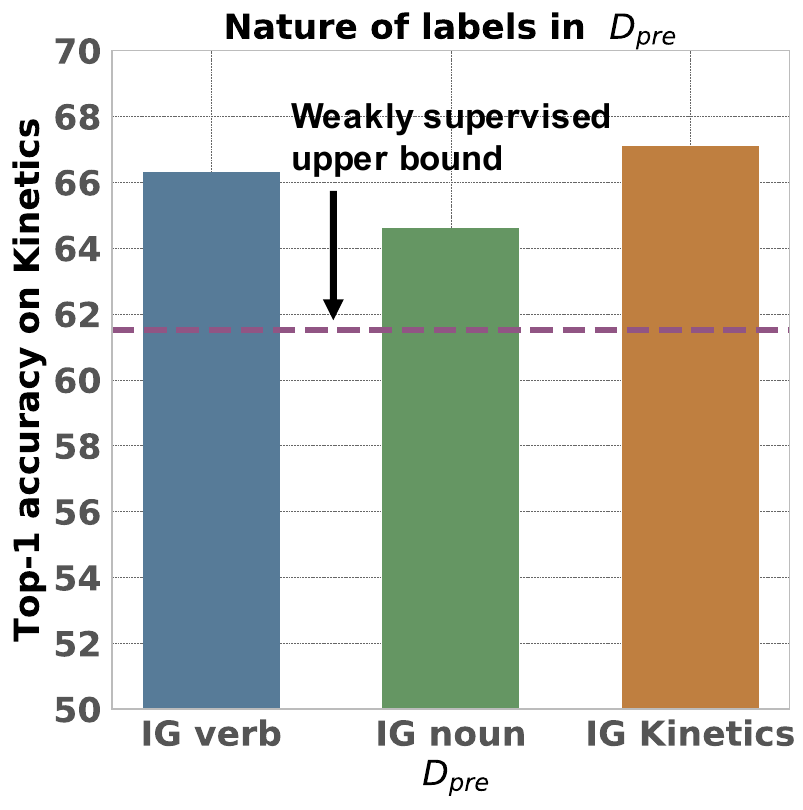} &
\includegraphics[width=0.24\textwidth]{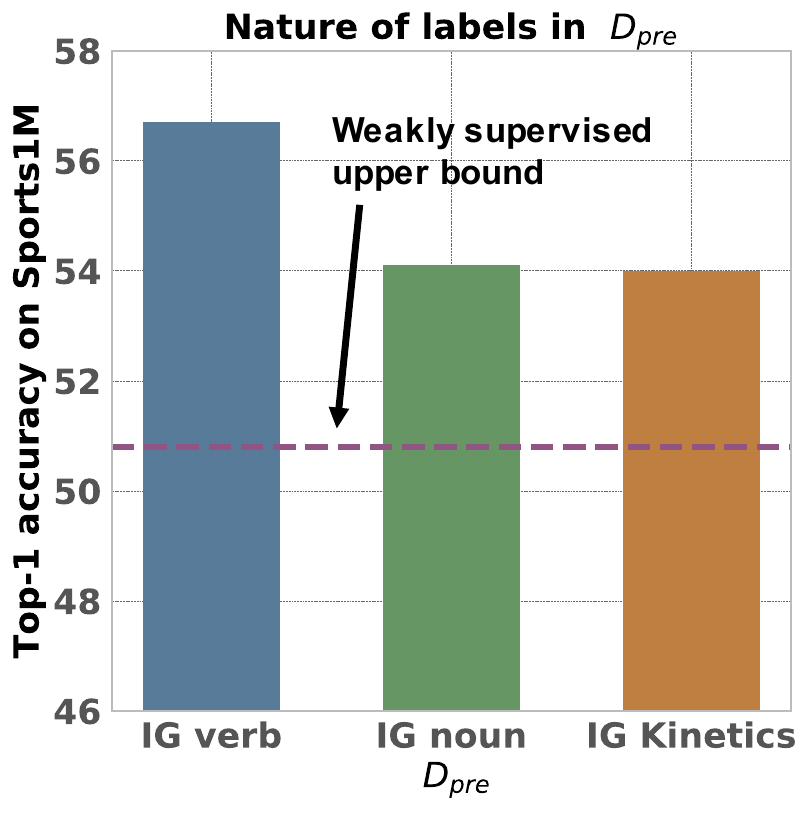} \\
\end{tabular}
\vspace{-0.2in}
\captionof{figure}{\textbf{Nature of pre-training labels ~(\cref{sec:analysis}):}} 
\vspace{-0.1in}
\label{fig:pretrained_properties_a}
\end{table}
 Figure~\ref{fig:pretrained_properties_a} shows the transfer learning performance on \kinetics and \sportsfull, where \clusterData is fixed to \igverball. For a fair comparison, we report the performance when \pretrainNet is pre-trained on \igverball, without applying \algorithmName (dotted magenta line). We make the following observations: first, all three \pretrainData datasets show significant improvements over the weakly-supervised upper bound upon applying \algorithmName, further reaffirming its generalizability. Second, \pretrainData $=$ \igkineticsuru yields a slightly higher performance on \kinetics, which indicates that prior domain knowledge of the downstream target task can help design \pretrainData for maximum benefit (as observed in~\cite{mahajan2018exploring, ghadiyaram2019large}).

\end{document}